\documentclass[10pt,twocolumn,letterpaper]{article}

\usepackage{cvpr}
\usepackage{times}
\usepackage{epsfig}
\usepackage{graphicx}
\usepackage{amsmath}
\usepackage{amssymb}
\usepackage{algorithm}
\usepackage{algpseudocode}
\usepackage{subfigure}
\usepackage{multirow} 
\cvprfinalcopy
\usepackage[pagebackref=true,breaklinks=true,letterpaper=true,colorlinks,bookmarks=false]{hyperref}

\def\cvprPaperID{2649} 

\ifcvprfinal\fi
\begin{document}

\title{Better Understanding Hierarchical Visual Relationship\\ for Image Caption}

\author{Zheng-cong Fei\textsuperscript{1,2}\\
\textsuperscript{1}Key Laboratory of Intelligent Information Processing of Chinese Academy of Sciences (CAS),\\
Institute of Computing Technology, CAS, Beijing, 100190, China\\
\textsuperscript{2}University of Chinese Academy of Sciences, Beijing, 100049, China\\
{\tt\small feizhengcong@ict.ac.cn}
}

\maketitle

\begin{abstract}
The Convolutional Neural Network (CNN) has been the dominant image feature extractor in computer vision for years. However, it fails to get the relationship between images/objects and their hierarchical interactions which can be helpful for representing and describing an image. In this paper, we propose a new design for image caption under a general encoder-decoder framework. It takes into account the hierarchical interactions between different abstraction levels of visual information in the images and their bounding-boxes. Specifically, we present CNN plus Graph Convolutional Network (GCN) architecture that novelly integrates both semantic and spatial visual relationships into image encoder.  The representations of regions in an image and the connections between images are refined by leveraging graph structure through GCN. With the learned multi-level features, our model capitalizes on the Transformer-based decoder for description generation. We conduct experiments on the COCO image captioning dataset. Evaluations show that our proposed model outperforms the previous state-of-the-art models in the task of image caption, leading to a better performance in terms of all evaluation metrics.

\end{abstract}

\section{Introduction}
Describing the content observed in an image, referred to  \emph{image caption}, has received a significant amount of attention in recent years. Image caption is an important task in its applications in various scenarios, \emph{e.g.}, recommendation in editing applications, usage in virtual assistants, image indexing, clustering in social media platforms and support of the disabled \cite{Hossain2018A}. 

Motivated by the recent advances in neural machine translation, current image captioning approaches typically follow an encoder-decoder framework \cite{Donahue2015Long,Ranzato2016Sequence,Vinyals2015Show,Xu2015Show}, which consists of a convolutional neural network-based image encoder and a recurrent neural network (RNN) based sentence decoder, with various variants for image captioning \cite{Fang2015From, Wu2015What}. Understanding an image largely depends on obtaining image features.
Despite the good performance of CNN as image feature extractor in different versions of these frameworks, they ignore the \emph{visual relationship} in the observed images.

\begin{figure}[t]
	\begin{center}
	{
			\includegraphics[width=0.95\linewidth]{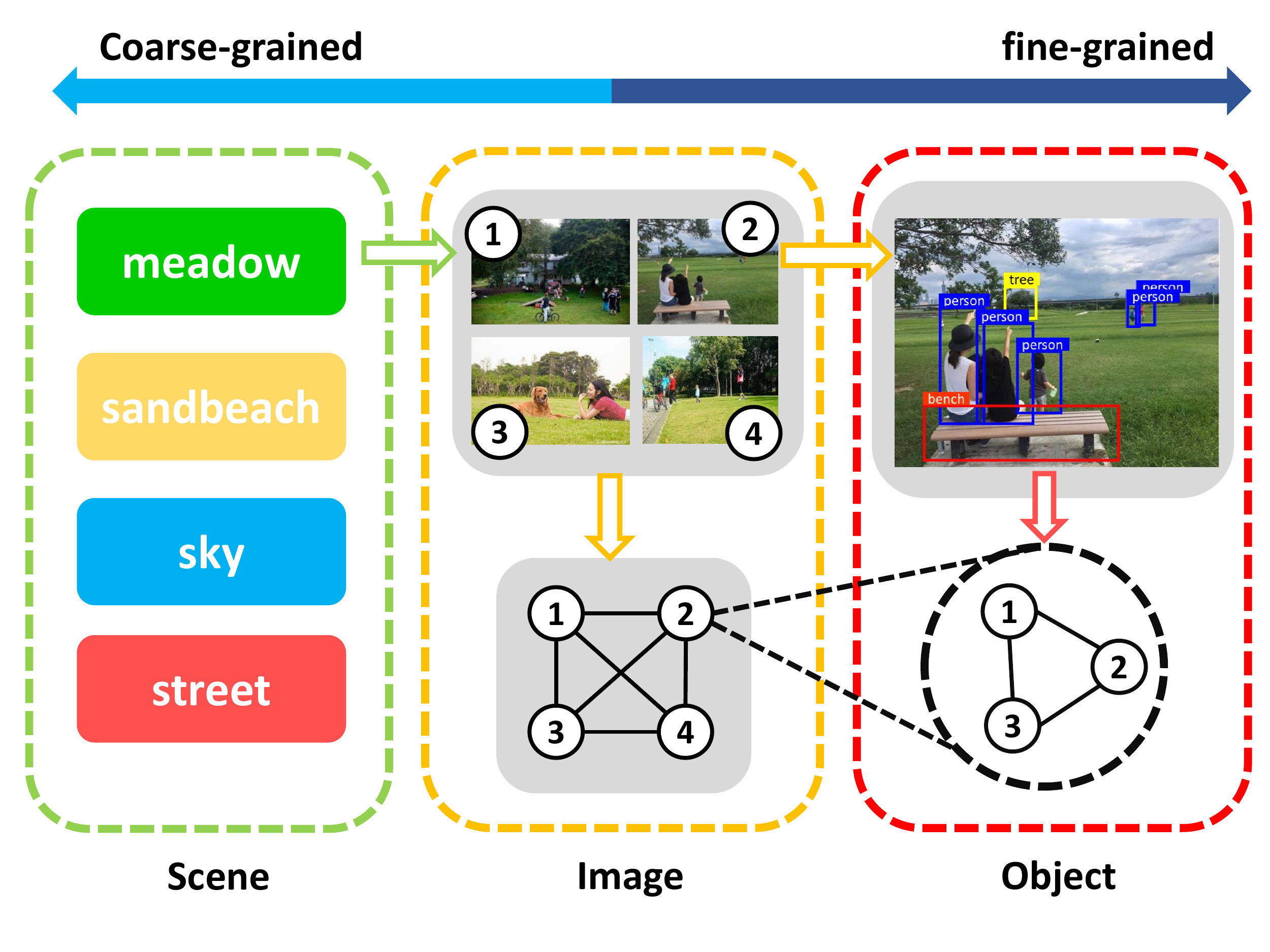}
		}
	\end{center}
	\caption{Current computer vision systems are incapable to provide an accurate caption over and above convolutions. To fill this gap, we proposed a visual hierarchical context-understanding deep learning architecture capable of adding visual and contextual information to the images’ objects. Here we illustrate an example of our proposed hierarchical interaction, considering different context levels, \emph{i.e.} objects, images, and scenes. Images are from the dataset available in \cite{Karpathy2016Deep}.}
	\label{fig1}
\end{figure}

\begin{figure*}[t]
	\begin{center}
		{
			\includegraphics[width=1\linewidth]{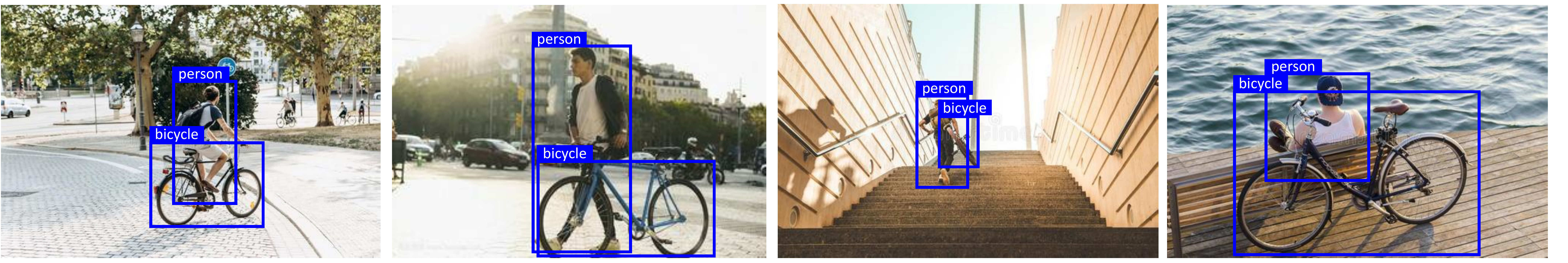}
		}
	\end{center}
	\caption{ Examples of images in which the same bounding-boxes’ concepts appear in different image contexts. Here we see that different semantic contexts present the same classes for the bounding-boxes, such as: “person” and “bicycle”. The visual relationships between these two objects in four images can be described as “riding”, “pushing”, “lifting” and “backing”, respectively. This scenario brings ambiguity and compromises the classifications of traditional end-to-end CNN architectures. Images used in this example are from the public dataset provided in \cite{Krizhevsky2012ImageNet}. }
	\label{fig2}
\end{figure*}

Visual relationship is the interactions or relative positions between objects detected in an image \cite{Yao2018Exploring}. The identification of visual relationships involves not only localizing and recognizing objects but also classifying the interaction between each pair of objects \cite{Li2017Scene,Lu2016Visual}. 
It is well believed that modeling relationships between objects would be helpful for representing and eventually describing an image.
Under such scenarios, traditional end-to-end CNN architectures collapse in relationship extracting. Because traditional CNNs ignore the semantic and spatial relations among objects to define the image context.
Figure \ref{fig2} illustrates a typical case. It is clear to see that the four imagines present a considerable overlap regarding their classes of bounding-boxes.
However, the visual relationship between object “person” and “bicycle” is intuitively different and each description sentence for the corresponding image should be differentiated as well. Only using the bounding-boxes’ classes to perform image caption will harm the final prediction, even in conjunction with other features from the image. This simple example testifies that state-of-the-art CNNs cannot cope with this issue, because different contexts can partially share the same objects. 

Hence, we believe that we need to grasp and describe the visual relations among objects to define the context of the image effectively.
Traditional convolution layers are incapable to cover this task because they are based on feature maps and cover the eyes to the relations between objects. Our insight is that, visual relationships between objects can be obtained through their semantic and spatial interactions. To construct this semantic connection, our method fuses the CNNs and Graph Convolutional Networks. 
On the other hand, common wisdom is that if two images are similar in terms of the same scene, the objects in both images are more likely to have similar actions. Therefore, we focus on the proposal of a visual hierarchical context-understanding architecture capable of defining the context of an image in different abstract levels.

Figure \ref{fig1} illustrates a problem with different context levels (coarse to fine-grained). For instance, a coarse-grained information is related to superclasses, such as “meadow”, “sand beach”, “sky”, “street”, etc. Each superclass is composed of different subclasses of images (medium-grained level). The fine-grained level considers the bounding-boxes (e.g. “person”, “tree” and “bench” belonging to the subclass “meadow”). To describe the global context of the image, our framework can build different graph representations for each image level. Each graph node encodes intrinsic information from the image or a bounding-box.
Our approach uses the fine and coarse-grained levels to build, respectively, inner (bounding-boxes of an image) and outer (superclass node linked to its subclasses nodes) interactions between nodes from the graph.  Using this structure, we can better grasp and describe the context in a hierarchical way. This occurs because we capture how images and their objects (bounding-boxes) interact with each other. Our framework can cope with different levels of granularity. Despite efforts regarding Graph Convolutional Network to capture object interactions in the image \cite{Yao2018Exploring}, to the best of our knowledge, previous literature differs from our proposal.

Our contributions are summarized as follows:
\begin{itemize}
    \item we developed a new visual hierarchical relationship framework which is capable of defining the context through a hierarchical scheme and considers different (scene, image, object) levels. Our proposed model takes into account not only the interactions between the bounding-boxes and their intrinsic characteristics but also the connections between the images from a similar context.
\item  We present a Transformer-based generation model, which does not rely on the RNN model and can focus on effective visual information to generate sentences. This structure is inherent, can be trained in parallel and shows comparable performance to other RNN-based methods on stand metrics. 
\end{itemize}



\section{Related Works}

\subsection{Image Feature Extraction}
CNN based deep learning approaches have performed successfully in image feature extraction tasks; however, CNNs do not work well with non-Euclidean features which is prevalent in many real-world applications \cite{Hossain2018A, Xiaochen2019}.
In this end, Graph Convolutional Network (GCN) was proposed to define convolutions on the non-grid structures by \cite{Tan2015Learning, Kipf2016Semi}. Thanks to its effectiveness, GCNs have been successfully applied to several research areas in computer vision, such as skeleton-based
action recognition \cite{Yan2018Spatial}, person re-identification \cite{Shen2018Person}, and video classification \cite{Wang2018Videos}.

Our framework is inspired in previous works \cite{Kipf2016Semi, Chen2017HARP, Yao2018Exploring} that try to grasp the interactions and the structure of a graph’s nodes through neural networks. Work of \cite{Chen2017HARP} proposes to learn low dimensional embeddings of a graph’s nodes. To do so, they decompose a graph on several levels (coarse to
fine-grained) and preserve the graph's structural features. In \cite{Kipf2016Semi}, the authors proposed a scalable learning method to convolutional neural networks on graphs. The work consists of using multiple graph convolution layers that promote a neighborhood aggregation of information. After $L$ layers, a given node fuses the information from its neighbors that are L-hops of distance in the graph. 
In summary, GCN can modify the feature vector, performing a kind of feature propagation, by aggregating more and more information at each hidden layer.

\subsection{Image Caption Generation}

In recent years, the task to generate natural sentences based on images has been widely studied \cite{Vinyals2016Show,Karpathy2016Deep}. Early researches take advantage of sentence template and heavily hand-designed systems \cite{Kulkarni2013Babytalk,Mitchell2012Midge}, which limits the application so far as to cause sensitivity to disturbance.
\cite{Vinyals2015Show} firstly proposed an encoder-decoder framework, which used the CNN as the image encoder and the RNN as the sentence decoder. Further, various improvement methods have been developed.  \cite{Xu2016Guiding} used
the semantic information to guide the LSTM along the sentence generation. \cite{Xu2015Show} proposed a spatial attention mechanism to attend to different parts of the image dynamically. \cite{Yang2016Encode} proposed a review network to extend the existing encoder-decoder models. \cite{Wu2015What,Yao2016Boosting} fed the attribute features into RNNs to leverage the high-level attributes. \cite{Anderson2017Bottom} proposed a combined bottom-up and top-down attention mechanism based on the object detection methods to generate descriptions. \cite{Yao2018Exploring} first introduced graph convolution networks to explore the visual relationship between objects to boost the image caption. As to our formulation, GCN is utilized to leverage different levels of visual information and then the Transformer is applied to decide which parts to follow when generating sentences.

\begin{figure*}[t]
	\begin{center}
		{\rule{0pt}{2in}
			\includegraphics[width=0.9\linewidth]{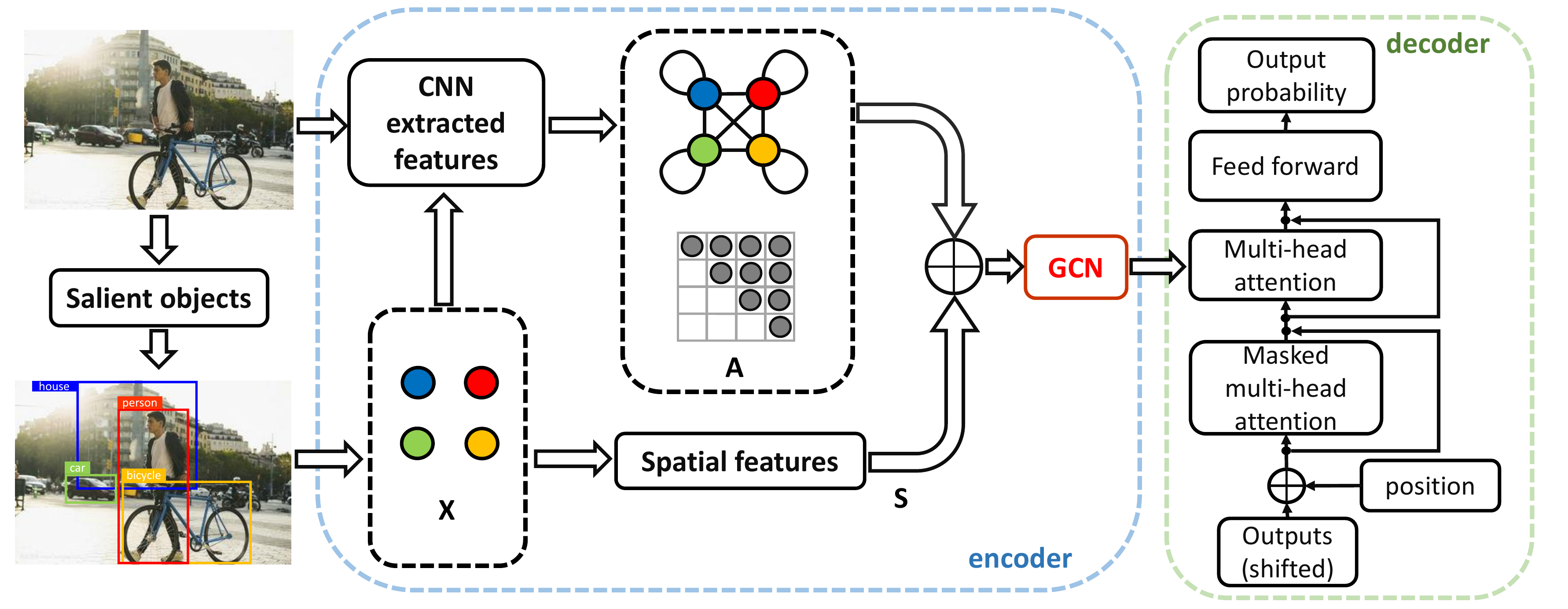}
		}
	\end{center}
	\caption{An overview of our Graph Convolutional Networks plus Transformer (GCN-T) for image captioning. GCN-T deals with different image levels. Here we see the representation of the object level. Faster R-CNN is first leveraged to detect a set of salient image regions. The bounding-boxes from the images are then described by a visual relation detection (visual relation classifier). For each image, its bounding-boxes will be nodes in a complete graph (\emph{i.e.} they interact
		with each other). An adjacency matrix is built and encodes the relationships between the bounding-boxes. Then, a GCN joins the visual and spatial features from each node with its interactions, propagating the information and resulting in our global image context. After that, the learned relation-aware region-level features from the graph are feed into one multi-head attention Transformer decoder for sentence generation.}
	\label{fig3}
\end{figure*}

\section{Methods}

In this section, we detail our GCN-T architecture which integrates hierarchical image structure into the encoder, pursuing a thorough image understanding to facilitate image captioning. 
Our model firstly utilizes an object detection module (\emph{e.g.}, Faster R-CNN \cite{Ren2017Faster}) to detect objects within images, aiming for generating a set of salient image regions containing objects. Semantic and spatial relationships between bounding-boxes and images are then defined through visual features and a graph structure.
Next, a GCN module in image encoder is leveraged to contextually refine the representation of each image region, resulting in relation-aware region representations. All of the encoded relation-aware region representations are further injected into the Transformer-based captioning framework, enabling multi-head attention mechanism for sentence generation.
Figure \ref{fig3} illustrates the representation of the object-level workflow (subclasses).

\subsection{Overview}

\textbf{Problem Formulation.}  The target of image captioning task is to describe a given image $\mathcal{I}$ with a textual sentence $\mathcal{S}$. Note that  the textual sentence $\mathcal{S} = \{w_1, w_2, \ldots, w_{N_s}\}$ is a word sequence consisting of $N_s$ words.  $w_t \in \mathbb{R}^{D_s} (1 \leq t \leq N_s) $ denote the $D_s$-dimensional textual feature of the $t^{th}$ word in the sentence $\mathcal{S}$. 

\textbf{Notation.} Objection detection model is first employed to produce the set of detected objects $\mathcal{V} = \{v_1,v_2,\ldots,v_K\} $ with $K$ image regions of objects in image $\mathcal{I}$ and $v_i \in \mathbb{R}^{D_v} (1 \leq t \leq K) $ denotes the $D_v$-dimensional feature of each image region.
Let $\mathcal{G}=(\mathcal{V},\mathcal{E})$ be a graph of $K$ nodes with object nodes $v_i \in \mathcal{V}$ and edge $e_{ij}=(v_i,v_j) \in \mathcal{E}$. Furthermore, let $\mathcal{A} \in \mathbb{R}^{K \times K}$ be the adjacency matrix associated with $\mathcal{G}$. Here, we seek to exploit graphs  $\mathcal{G}=(\mathcal{V},\mathcal{E})$ to contextually refine the representation of each image region, which is endowed with the inherent visual relationships between objects and images.


\subsection{Visual Relationship Extraction}
In this section, we mainly introduce the extraction of both the semantic relationship and spatial features between objects.
For each image $\mathcal{I}$, we use pre-trained Faster R-CNN to extract salient object features $\mathcal{V}$. Specifically, we extract the image features from the final convolutional layer of CNN and use spatial adaptive average pooling to resize the features to a fixed-size spatial object representation.

\subsubsection{Semantic Visual Relationship}

Following \cite{Yao2018Exploring}, we simplify the detection as a classification task to learn semantic relation classifier on visual relationship
benchmarks \cite{Krishna2017Visual}. The general meaning of the semantic relationship is an action or interaction between one object and another object. Hence, the relationship can be expressed as $<subject-predicate-object>$. Given two detected regions of $v_i$ and $v_j$ with an image $\mathcal{I}$, we devise a simple deep classification model to predict the semantic relation between $v_i$ and $v_j$ depending on the union bounding box which covers the two objects together,
\begin{equation}
	r_{semantic}=\mathcal{F}(v_i,v_j)
	\label{eq0}
\end{equation}
where $r_{semantic}$ is a softmax probability over semantic relation classes plus a non-relation class.

\subsubsection{Spatial Visual Relationship}
we consider the spatial features \cite{Peyre2017Weakly} from each pair of bounding-boxes $v_i^p=[\alpha_i, \beta_i, \gamma_i, \delta_i]$ and $v_j^p=[\alpha_j, \beta_j, \gamma_j, \delta_j]$, where $(\alpha,\beta)$ are coordinates of the center of the box and $(\gamma,\delta)$ are the width and height of the box. Through the Equation \ref{eq-1}, we obtained a 6-dimensional spatial feature vector.
\begin{equation}
	r(v_i^p,v_j^p)=[\dfrac{\alpha_j-\alpha_i}{\sqrt{\gamma_i\delta_i}},\dfrac{\beta_j-\beta_i}{\sqrt{\gamma_i\delta_i}},\sqrt{\frac{\gamma_j \delta_j}{\gamma_i \delta_i}},\dfrac{v_i \cap v_j}{v_i \cup v_j},\dfrac{\gamma_i}{\delta_i},\dfrac{\gamma_j}{\delta_j} ]
	\label{eq-1}
\end{equation}
The first two features represent the renormalized translation between the two boxes; the third is the ratio of box sizes; the fourth is the overlap between boxes, and the fifth and sixth encode the aspect ratio of each box, respectively.

To obtain a well-suited representation, we perform the discretization of the feature vector into $m$ bins. For this, the spatial configurations $r(v_i^p,v_j^p)$ are generated by a mixture of $m$ Gaussians and the parameters of the Gaussian Mixture Model are fitted to the training pairs of boxes. We used as
our spatial features the scores that represent the probability of assignment to each of the $m$ clusters. 

\subsection{Hierarchical Graph Construction}

Each bounding-box from an input image $\mathcal{I}$ will be a node in a graph (interact with each other). According to the previous outcome, our method builds a graph  $\mathcal{G}=(\mathcal{V},\mathcal{E})$, where $\mathcal{V}$ and $\mathcal{E}$ represent the node set and the edge set, respectively. Next, it builds a complete graph to promote the information flow among all bounding-boxes, even those far from each other (and therefore not considered in a given receptive field). Then, the method creates an adjacency matrix $\mathcal{A}$ to encode the relationships between the bounding-boxes. In the last phase, a GCN joins the features (visual and spatial) to propagate the information from each node.

Extensions would be applied to our framework, considering other types of features. We explored the inclusion of spatial features (\emph{e.g.} the normalized translation between the bounding-boxes, the ratio of box sizes, among others). To better describe these spatial features, we expanded them to a fine-grained representation through a Gaussian Mixture Model (GMM) discretization.
Considering this extension, we proposed a propagation rule to fuse visual and complementary features (e.g. spatial) into a GCN, and it is formally defined by Equation \ref{eq1}.
\begin{equation}
\mathcal{H}^{(l+1)}=\sigma(\hat{\mathcal{A}}\mathcal{H}^{(l)}S^{(l)}W^{(l)})
\label{eq1}
\end{equation}
where $\hat{\mathcal{A}}$ is the renormalized adjacency matrix with added self-loops; $\mathcal{H}^{(0)}$ represents the input visual features for each node; and $S^{(0)}$ denotes the initial complementary features
(\emph{i.e.} describes each node or each pair of nodes in the graph).

From Equation \ref{eq1}, it is clear to note that our method can fuse different types of features (\emph{e.g.} visual and spatial) to describe a node from the graph. It can also consider separated visual and spatial features (\emph{i.e.} w/o fusion) or other combinations. Moreover, our framework can be straightforwardly extended to accommodate other policies. Our method is easily generalized to other types of information (\emph{e.g.} obtained from graph structure, among others) to generate different propagation rules. 

The information related to the context is intrinsically detected by GCN-T through the representations of the nodes joined with our graph construction. It considers not only each object in a given image and its interactions but also the connection among images from the same context. This provides the hierarchical description and interaction, resulting in our global context.
To the best of our knowledge, this is the first time that GCN is used to capture the global context (\emph{i.e.} class) of an image through the interaction between its objects and connections between images from the same context. For instance, once the bounding-boxes from an image (of a given subclass) are nodes in a graph structure, GCN-T can consider each image in a given superclass as nodes in a supergraph that contains a subgraph of bounding-boxes nodes
(see Figure \ref{fig1}).

\subsection{Attention in Transformer}
Instead of applying RNN or LSTM, we introduce the Transformer model to the description sentence generation in this task. Transformer can reach significantly better performances in many tasks, such as machine translation \cite{Ott2018Scaling}, question answering \cite{Yang2018HotpotQA}, and natural language generation \cite{Liu2018Generating}. In our preliminary experiments, we have found that the Transformer can also achieve better performance in this task, and thus we apply this model to visual image caption.

The Transformer decoder consists of an embedding layer and multiple decoder layers. Each decoder layer has a self-attention module and a Point-Wis Feed-Forward Network (FFN).  Moreover, each of them contains a multi-head context
attention to extract information from the source site context. The decoder generates a representation $o$ at each decoding time step.

In order to predict a word at each decoding time step, the top layer of the decoder, namely the output layer, generates a probability distribution over the vocabulary,
\begin{equation}
P_w=\text{softmax}(W_To+b_T)
\end{equation}
where $W_T \in \mathbb{R}^{|V|\times d}$ and $b_T \in \mathbb{R}^{|V|}$ are weight and bias parameters, and $V$ refers to the vocabulary.

In practice, different from the machine translation task, we need to combine the image with the text information. At the first decoder layer, all the inputs are identical. That means that the keys, values, and queries are the same matrices, and this mechanism is called \emph{self-attention}. This controls the relationship between the whole sequence. At the second decoder layer, the keys and the values are the matrices generated by the GCN, which reserves the spatial and semantic image information. The queries are the outputs by the first decoder layer, which means the sentence information. The target of this attention is to make the relation between the information of the image and the sentence information.

\subsection{Training and Inference}

\begin{algorithm}[t]
	\caption{Hierarchical GCN Training}
	\label{a1}
	\begin{algorithmic}[1]
		\Require
		A set of images $\mathbb{I}$ and their respective object bounding-boxes $\mathbb{V}$.
		\Ensure
		Trained graph convolutional network model.
		\State X $\mapsto$ $\emptyset$;
		\State \textbf{if} level = object-class \textbf{then}
		\State \qquad X $\mapsto$ visual features of the bounding-boxes using CNNs;
		\State \qquad S $\mapsto$ spatial features;
		\State \qquad \textbf{for} each image $\mathcal{I}$ in image set $\mathbb{I}$ \textbf{do}
		\State \qquad \qquad $\mathcal{V}_i$ is composed of object bounding-boxes;
		\State \qquad \qquad build a complete graph $\mathcal{G}_i=(\mathcal{V}_i,\mathcal{E}_i)$;
		\State \qquad \textbf{end}
		\State \textbf{end}
		\State \textbf{else if} level = image-class \textbf{then}
		\State \qquad X $\mapsto$ visual features of the images using CNNs;
		\State \qquad \textbf{for} each image $\mathcal{I}$ in image set $\mathbb{I}$ \textbf{do}
		\State \qquad \qquad $\mathcal{V}_j$ is composed of images from a similar context $j$;
		\State \qquad \qquad build a complete graph $\mathcal{G}_j=(\mathcal{V}_j,\mathcal{E}_j)$;
		\State \qquad \textbf{end}
		\State \textbf{end}
		\State \textbf{else if}  level = hierarchical \textbf{then}
		\State \qquad X $\mapsto$ visual features of the images and the object bounding-boxes using CNNs;
		\State \qquad S $\mapsto$ spatial features;
		\State \qquad \textbf{for} each image $\mathcal{I}$ in image set $\mathbb{I}$ \textbf{do}
		\State \qquad \qquad $\mathcal{V}_j$ is composed of images from a similar context $j$ and their bounding-boxes;
		\State \qquad \qquad build a complete graph $\mathcal{G}_j=(\mathcal{V}_j,\mathcal{E}_j)$;
		\State \qquad \textbf{end}
		\State compute the adjacency matrix $\mathcal{A}$ from $\mathcal{G}$;
		\State \textbf{repeat}	
		\State \qquad backpropagate and optimize parameters $\{W\}$;
		\State \textbf{until} $\{W\}$ has coverged;	
		
		\\
		\Return trained model;
	\end{algorithmic}
\end{algorithm}

The total training process can be described as follows. Firstly, an image $\mathcal{I}$  will be sent to the CNN plus GCN model, then we will get the extracted feature. The feature matrix is sent to the Transformer as the second sub-layer input. The ground-truth sentence embedding matrix is sent as the Transformer input. At the last, the model gets the probability distribution $p(\mathcal{S}'|\mathcal{S}, \mathcal{I})$ for the image, where $\mathcal{S} \in R^{N_s \times D_s}$ is the $N_s$ length ground-truth sentence embedding matrix and $\mathcal{S}' \in R^{N_s \times D_s}$ is the sentence generated by the model which shift right relative to the $\mathcal{S}$. The Transformer now gets the whole sentence probability for the current image.


To learn this model, we use the supervised learning method. Given the target ground truth sequence $\mathcal{S} = \{w_1, w_2, \ldots, w_{N_s}\}$, the model would be trained by minimizing the cross-entropy loss (XE) which is the same as that described in \cite{Vinyals2016Show}. It is shown as follows,
\begin{equation}
	logp(\mathcal{S}|\mathcal{I})=\sum_{t=0}^{N_s} logp(w_t|\mathcal{I},w_0, \ldots, w_{t-1}; \theta)
	\label{eq6}
\end{equation}
where $\theta$ is the parameter of the model and $(\mathcal{S}, \mathcal{I})$ is the training example pair. We optimize the sum of the log probabilities as described in the above over the whole training set. In addition, Algorithm \ref{a1} details the training procedure of hierarchical graph convolutional network. 

The inference is similar to the general encoder-decoder framework, and the word will be generated one by one at a time. Firstly, we also need to begin with the start token $<S>$, and generate the first word by $p(w_1|w_0, \mathcal{I})$.
Afterwards, we get the dictionary probability $w_1 \sim p(w_1|w_0, \mathcal{I})$ at the first time. We can use the greedy method or the beam search method to select the possible word. Then, $w_1$ is fed back into the network to generate the following word $w_2$. This process will be continued until the end token $<E>$ is
reached.

\section{Experiments}

We evaluated the performance of our proposed GCN-T model on the COCO captioning dataset (COCO) \cite{Lin2014Microsoft} for image captioning task. In addition, Visual Genome \cite{Karpathy2016Deep} is utilized to pre-train the object detector and MIT67 Dataset \cite{Quattoni2001Recognizing} is utilized to train hierarchical GCN with Algorithm \ref{a1} in our GCN-T model. 

\subsection{Datasets and Experimental Settings}

\begin{table*}[!h]
	\begin{center}
		\begin{tabular}{|l|c|c|c|c|c|c|c|c|c|c|}
			\hline
			     & \multicolumn{5}{|c|}{\textbf{Cross-Entropy Loss}} &\multicolumn{5}{|c|}{\textbf{CIDEr-D Score Optimization}} \\ \cline{2-11}
			    &B@4&M&R&C&S&B@4&M&R&C&S\\
			\hline\hline
			LSTM \cite{Vinyals2016Show} &29.6&25.2&52.6&94.0& - &31.9&25.5&54.3&106.3& - \\
			SCST \cite{Rennie2017Self} &30.0&25.9&53.4&99.4&-&34.2&26.7&55.7&114.0& - \\
			ADP-ATT \cite{Lu2016Knowing} &33.2&26.6&-&108.5&-&-&-&-&-&- \\
			LSTM-A \cite{Yao2016Boosting} &35.2&26.9&55.8&108.8&20.0&35.5&27.3&56.8&118.3&20.8\\
			Up-Down \cite{Anderson2017Bottom}  &36.2&27.0&56.4&113.5&20.3&36.3&27.7&56.9&120.1&21.4\\ \hline
			GCN-LSTM \cite{Yao2018Exploring} &37.0&28.1&57.1&117.1&21.1&38.3&28.6&58.1&128.7&22.1\\
			GCN-T$_{obj}$  &37.2&28.5&57.3&118.8&21.5&38.6&29.0&58.7&128.8&22.5\\
			GCN-T&\textbf{37.8}&\textbf{28.8}&\textbf{57.7}&\textbf{119.5}&\textbf{21.5}&\textbf{38.9}&\textbf{29.1}&\textbf{59.4}&\textbf{129.7}&\textbf{22.8}\\
			\hline
		\end{tabular}
	\end{center}
	\caption{Performance of our GCN-T and other state-of-the-art methods on COCO, where B@4, M, R, C and S are short for BLEU@4, METEOR, ROUGE-L,
		CIDEr-D and SPICE scores. All values are reported as percentage (\%).}
	\label{tab1}
\end{table*}

\textbf{COCO} is a standard benchmark for image captioning task which contains 123,287 images (82,783 for training and 40,504 for validation) and each image is annotated with 5 descriptions by humans. Since the annotated descriptions of the official testing set are not provided, we utilize Karpathy split (113,287 for training, 5,000 for validation and 5,000 for testing) as in \cite{Anderson2017Bottom}. According to \cite{Karpathy2016Deep}, all the training sentences are converted to lower case and we omit rare words which occur less than 5 times. Therefore, the final vocabulary includes 10,201 unique words.

\textbf{Visual Genome}, which contains images with annotated objects, attributes, and relationships, is adopted to train Faster R-CNN for object detection. In this paper, we follow the setting in \cite{Anderson2017Bottom, Yao2018Exploring} and take 98,077 images for training, 5,000 for validation, and 5,000 for testing. As in \cite{Anderson2017Bottom}, 1,600 objects and 400 attributes are considered from Visual Genome for training Faster R-CNN with two branches for predicting objects and attribute classes.

\textbf{MIT67 Dataset} contains 67 classes (subclasses) from indoor scenes covering a wide range of 5 contexts (superclasses), including “leisure”, “working place”, “home”, “store” and “public space” scene categories \cite{Quattoni2001Recognizing}. From this dataset, we can explicitly train our hierarchical level strategy which considers superclasses and subclasses. Then, besides considering the bounding-boxes in an image as nodes in a graph structure, we can also regard each image in a given superclass as nodes in a supergraph. It is noted that our GCN-T model can reach an accuracy of almost 65.8\% at a hierarchical level.

\textbf{Implementation Details.} For each image, we apply Faster R-CNN to detect objects within this image and select top $K$ = 36 regions with the highest detection confidences to represent the image. The dimension of each region $D_v$ is set as 2,048. For the Transformer model, we set the model size which is represented as $D_s$ to be 512. 
The captioning models with Hierarchical GCN are mainly implemented with PyTorch, optimized with Adam \cite{Kingma2014Adam}. We set the initial learning rate as 0.0005 and the mini-batch size as 1,024. The momentum and the weight-decay are 0.8 and 0.999 respectively.  The maximum training iteration is set as 35 epochs. At inference, beam search strategy is adopted and the beam size is set as 3.

\textbf{Evaluation Metrics.} According to \cite{Yao2018Exploring}, We select five types of metrics: BLEU@N \cite{Papineni2002BLEU}, METEOR \cite{Lavie2007METEOR}, ROUGE-L \cite{Flick2004ROUGE}, CIDEr-D \cite{Vedantam2015CIDEr} and SPICE \cite{Anderson2016SPICE}.

\textbf{Compared Approaches.}  We compared the following state-of-the-art methods: (1) \textbf{LSTM} \cite{Vinyals2016Show} is the simplest CNN plus RNN model which only feeds image into LSTM at the initial time step. We directly extract results reported in \cite{Yao2018Exploring}. (2) \textbf{SCST} \cite{Rennie2017Self} employs a self-critical sequence training strategy to train a modified visual attention-based captioning model in \cite{Xu2015Show}  (3) \textbf{ADP-ATT} \cite{Lu2016Knowing} develops an adaptive attention-based encoder-decoder model for automatically determining whether to attend
to the image and which image regions to focus for caption. (4) \textbf{LSTM-A} \cite{Yao2016Boosting} integrates semantic attributes into CNN plus RNN captioning model for boosting image captioning. (5) \textbf{Up-Down} \cite{Anderson2017Bottom} designs a combined bottom-up and top-down attention mechanism that enables region-level attention to be calculated to boost image caption. (6) \textbf{GCN-LSTM} \cite{Yao2018Exploring} further exploits visual relationships between objects through graph convolutional networks. (7) \textbf{GCN-T} is the proposal in this paper. Moreover, one slightly different setting of GCN-T is named as \textbf{GCN-T}$_{obj}$ which is trained with only object-class visual relationship.

Note that for a fair comparison, all the baselines and our model adopt ResNet-101 as the basic architecture of image feature extractor. 
Moreover, results are reported for models optimized with both cross-entropy loss or expected sentence-level reward loss. The sentence-level reward is measured with the CIDEr-D score.

\subsection{Performance Comparison and Analysis}

\subsubsection{Quantitative Analysis}

The performances of different models on the COCO dataset for image captioning task are showed in Table \ref{tab1}. In general, the results across all metrics and two optimization methods (Cross-Entropy Loss and CIDEr-D Score Optimization) indicate that our proposed GCN-T exhibits best performances among
other approaches, including non-attention models (LSTM, LSTM-A), attention-based approaches (SCST, ADPATT and Up-Down)  and graph-based approach (GCN-LSTM, GCN-T$_{obj}$). In particular, the  GCN-T by integrating hierarchy context levels makes the absolute improvement over GCN-LSTM by 2.4\% and 0.8\% in terms of CIDEr-D and BLEU@4 optimized with cross-entropy loss, respectively, which is generally considered as significant progress on this benchmark.
The results generally highlight the key advantage of exploiting the hierarchal visual relationship in an image from the object level, image level, scene level, pursuing a better understanding of image captioning. On the other hand, by injecting the high-level semantic attributes into LSTM-based decoder, LSTM-A outperforms LSTM that trains decoder only depending on the input image. Nevertheless, the attention-based methods (SCST, ADP-ATT, and Up-Down) achieve better performance than LSTM-A, which verifies the advantage of attention mechanism that dynamically focuses on image regions for sentence generation. 
Furthermore, GCN-LSTM by exploring the relations between objects to enrich region-level features, improves SCST, ADP-ATTand UpDown. However, the performances of GCN-LSTM are lower than GCN-T that additionally exploits hierarchical visual relationships in an image for enhancing all the object-level, image-level, and scene-level features and eventually boosting image captioning. In addition, by optimizing the captioning models with CIDEr-D score instead of cross-entropy loss, the CIDEr-D score of GCN-T is further boosted up to 129.7\%. This confirms that the self-critical training strategy is an effective way to amend the discrepancy between training and inference, and improve sentence generation regardless of image captioning approaches.

\begin{figure*}[t]
	\begin{center}
		{\rule{0pt}{2in}
			\includegraphics[width=0.98\linewidth]{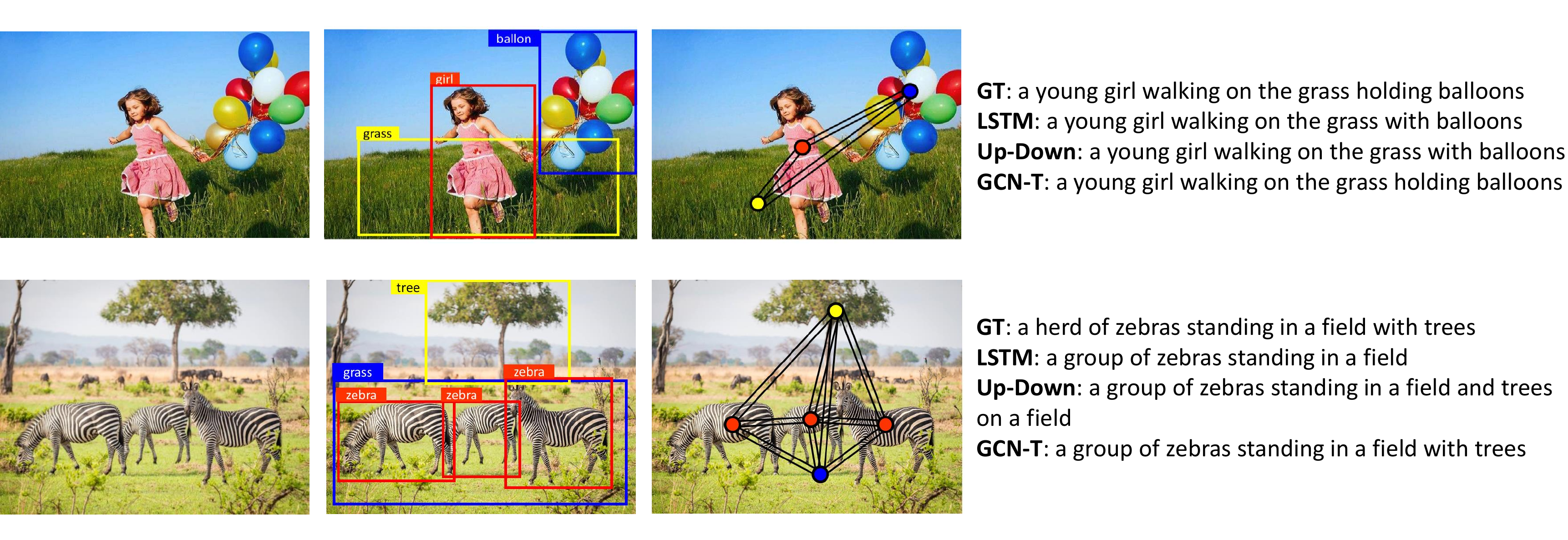}
		}
	\end{center}
	\caption{Two image examples from \cite{Lin2014Microsoft} with object regions, visual relationship, and sentence generation results. The output
		sentences are generated by (1) Ground Truth (GT): One ground truth sentence, (2) LSTM, (3) Up-Down and (4) our GCN-T.}
	\label{fig4}
\end{figure*}

\subsubsection{Qualitative Analysis}
Figure \ref{fig4} shows a few image examples with object regions, visual relationships, human-annotated ground truth sentences and captions generated by LSTM, Up-Down, and our proposed GCN-T, respectively. From these example results, it is clear that the three automatic methods can generate somewhat relevant and logically correct sentences,  while the output of the sentence by our model GCN-T is more descriptive. Since our model can enrich semantics with hierarchical visual relationships in graphs to boost image caption.
For example, compared to the same sentence segment "with balloons" in the sentences generated by LSTM and Up-Down for the first image, "holding balloons" in our proposed GCN-T depicts the image content more comprehensive and accurate, as the detected relation "holding" in graph is encoded into relation-aware features for guiding sentence generation.
The results again indicate the advantage of generating the sentence
with the help of a hierarchical visual relationship.

\subsubsection{Performance on COCO Online Testing Server}

\begin{table*}
	\begin{center}
		\begin{tabular}{|l|c|c|c|c|c|c|c|c|c|c|}
			\hline
			&\multicolumn{2}{|c|}{B@2}&\multicolumn{2}{|c|}{B@4}&\multicolumn{2}{|c|}{M}&\multicolumn{2}{|c|}{R} &\multicolumn{2}{|c|}{C}  \\
			\cline{2-11}
			&c5&c40&c5&c40&c5&c40&c5&c40&c5&c40\\
			\hline \hline
			GCN-T&\textbf{66.0}&\textbf{90.2}&\textbf{39.5}&\textbf{71.1}&\textbf{28.5}&\textbf{38.2}&\textbf{58.8}&\textbf{74.0}&\textbf{127.4}&\textbf{129.5}\\
			GCN-LSTM \cite{Yao2018Exploring}&65.5&89.2&38.7&69.7&28.5&37.6&58.5&73.4&125.3&126.5\\
			RFNet\cite{Jiang2018Recurrent}&64.9&89.3&38.0&69.2&28.2&37.2&58.2&73.1&122.9&125.1\\
			Up-Down\cite{Anderson2017Bottom}&64.1&88.7&36.9&68.5&27.6&36.7&57.1&72.4&117.8&120.5\\
			LSTM-A\cite{Yao2016Boosting}&62.7&86.7&35.4&65.2&27.0&35.5&56.3&70.7&114.7&116.7\\
			\hline
		\end{tabular}
	\end{center}
	\caption{Performance of the top-ranking published state-of-the-art image captioning models on the online COCO testing server, where B@N, M, R, and C are short for BLEU@N, METEOR, ROUGE-L, and CIDEr-D scores. All values are reported as percentage (\%).}
	\label{tab2}
\end{table*}

Besides, we submitted the run of GCN-T optimized with the CIDEr-D score to the online COCO testing server and evaluated the capacity on the official testing set. Table \ref{tab2} illustrates the performance of top-ranking image captioning models.
The latest top-ranking performing systems which have been officially published include GCN-LSTM, RFNet, Up-Down, and LSTM-A. However, it is significant that our proposed GCN-T model leads to better performance against all the other top-performing systems on the Leaderboard with most evaluation metrics both  5 reference captions and 40 reference captions.

\subsubsection{Human Perception Evaluation}
According to \cite{Yao2018Exploring}, since the automatic sentence evaluation metrics do not necessarily correlate with human judgment, we additionally conducted a human perception experiment to compare our GCN-T with three baselines, \emph{i.e.}, LSTM, Up-Down and GCN-LSTM. 

\textbf{Experiment setting.} We invite 24 evaluators and randomly select 1K images from testing set for human evaluation. All the evaluators are randomly grouped into two teams. We show the first team each image with four auto-generated sentences plus four human-annotated captions and ask them: Do the systems produce human-like sentences? Instead, we show the
second team only one description at a time, which may be generated by captioning methods or human annotation. The evaluators are required to answer: Can you distinguish human annotation from that by the caption system? 

\textbf{Evaluation Metrics.} Based on evaluators’ feedback and  evaluation metrics in \cite{Yao2018Exploring}, we select two metrics:
(1) M1: percentage of captions that are as well as or even better than human annotation; (2) M2: percentage of captions that pass the Turing
Test. 

\textbf{Result Analysis.} The results of M1 score of GCN-T, GCN-LSTM, Up-Down, and LSTM are 76.7\%, 74.1\%, 69.5\%, and 50.7\%, respectively. In terms of M2 score, Human, GCN-T, GCN-LSTM, Up-Down and LSTM achieve 92.4\%, 88.2\%, 81.5\%, 74.5\%, and 57.2\%. Above all, our GCN-T model achieves the best performance among other image caption models under both criteria.

\section{Conclutions}

In this paper, we introduce a novel framework for hierarchical visual relationship exploring. 
The proposed framework is capable of describing the global context of images under different scenarios to boost captioning. Besides the intrinsic features, to reach this prediction, our model grasps and describes the semantic relationships and spatial relationships among objects from an image (in fine-grained level) and interactions between images from a relative context (in a coarse-grained level). Furthermore, We utilize the Transformer which only uses stack attention layers to learn the sequence relationships among the language for sentence generation.  
The extensive experiments testify that our GCN-T model outperforms the traditional state-of-the-art architectures by a considerable margin.

{\small
\bibliographystyle{ieee_fullname}
\bibliography{cite}
}

\end{document}